\documentclass{article}

\usepackage{microtype}
\usepackage{graphicx}
\usepackage{subcaption}
\usepackage{booktabs}
\usepackage{hyperref}
\usepackage{amsmath}
\usepackage{amssymb}
\usepackage{mathtools}
\usepackage{amsthm}
\usepackage{dblfloatfix}
\usepackage{placeins}
\usepackage[capitalize,noabbrev]{cleveref}

\usepackage[accepted]{icml2026}

\icmltitlerunning{m2sv: Map-to-Street-View Spatial Reasoning Benchmark}

\begin{document}

\twocolumn[
\icmltitle{m2sv: A Scalable Benchmark for Map-to-Street-View Spatial Reasoning}

\icmlsetsymbol{equal}{*}

\begin{icmlauthorlist}
  \icmlauthor{Yosub Shin}{uhm}
  \icmlauthor{Michael Buriek}{uhm,pwc}
  \icmlauthor{Igor Molybog}{uhm}
  \end{icmlauthorlist}
  
  \icmlaffiliation{uhm}{University of Hawai'i at M\=anoa, Honolulu, HI, USA}
  \icmlaffiliation{pwc}{PwC, USA}
  
  \icmlcorrespondingauthor{Yosub Shin}{yosubs@hawaii.edu}
  
  \icmlkeywords{vision-language models, spatial reasoning, geospatial understanding, multimodal benchmarks, map-to-image alignment, grounded reasoning}

\vskip 0.3in
]

\printAffiliationsAndNotice{}

\begin{abstract}
Vision--language models (VLMs) achieve strong performance on many multimodal
benchmarks but remain brittle on spatial reasoning tasks that require aligning
abstract overhead representations with egocentric views. We introduce
\textbf{m2sv}, a scalable benchmark for map-to-street-view spatial reasoning that
asks models to infer camera viewing direction by aligning a north-up overhead map
with a Street View image captured at the same real-world intersection.
We release \textbf{m2sv-20k}, a geographically diverse benchmark with controlled
ambiguity, along with \textbf{m2sv-sft-11k}, a curated set of structured reasoning
traces for supervised fine-tuning.

Despite strong performance on existing multimodal benchmarks, the best evaluated
VLM achieves only 65.2\% accuracy on m2sv, below human annotators who reach
72.0\% on average (and 95\% for an expert) with strong inter-annotator agreement
($\kappa$ up to 0.76). While supervised fine-tuning and reinforcement learning
yield consistent gains, cross-benchmark evaluations reveal limited transfer.
Beyond aggregate accuracy, we systematically analyze \emph{difficulty} in
map-to-street-view reasoning using both structural signals and human effort, and
conduct an extensive failure analysis of adapted open models. Our findings
highlight persistent gaps in geometric alignment, evidence aggregation, and
reasoning consistency, motivating future work on grounded spatial reasoning
across viewpoints.
\end{abstract}

\begin{figure*}[!t]
  \centering
  \begin{minipage}[t]{0.28\linewidth}
    \vspace{0pt}
    \centering
    \begin{subfigure}[t]{\linewidth}
      \centering
      \includegraphics[width=\linewidth]{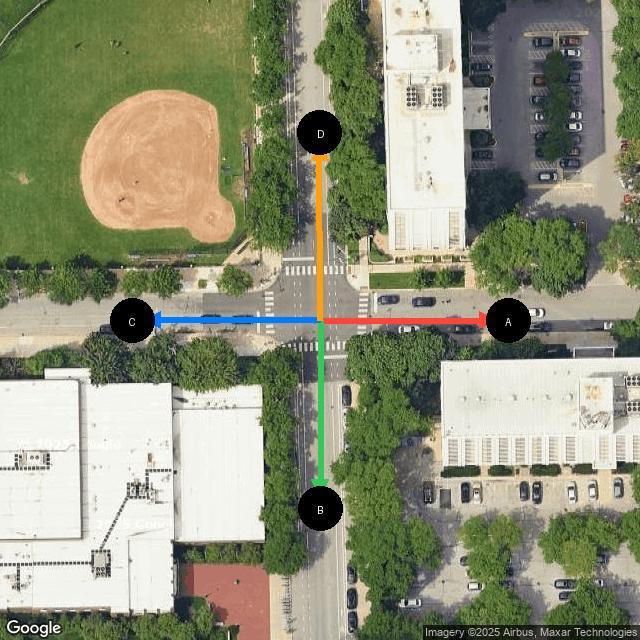}
      \caption{North-up overhead map with labeled candidate directions.}
      \label{fig:m2sv-example-map}
    \end{subfigure}

    \vspace{0.5em}
    \begin{subfigure}[t]{\linewidth}
      \centering
      \includegraphics[width=\linewidth]{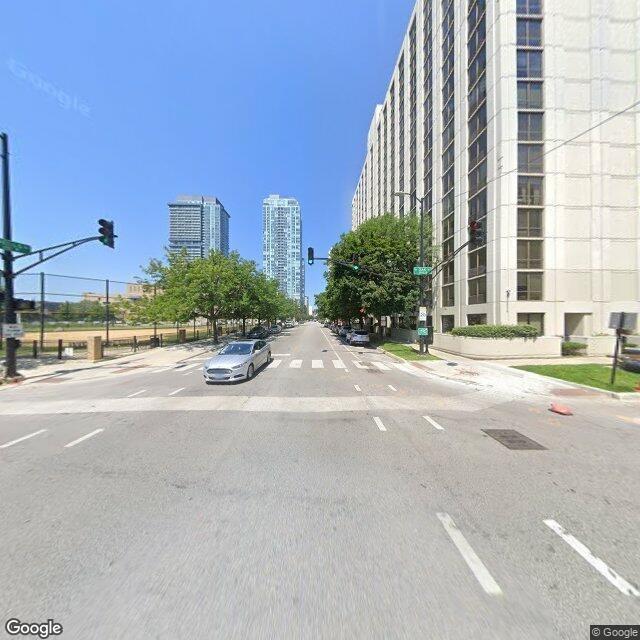}
      \caption{Street View image captured at the same intersection.}
      \label{fig:m2sv-example-sv}
    \end{subfigure}
  \end{minipage}\hfill
  \begin{minipage}[t]{0.71\linewidth}
    \vspace{0pt}
    \footnotesize
    \begin{tabular}{p{0.96\linewidth}}
      \toprule
      \textbf{Question} \\
      Which labeled direction on the map corresponds to the direction in which the street view photo was taken? Options: A, B, C, D. \\
      \midrule
      \textbf{Generated trace (Gemini-2.5-Pro)} \\
      \textbf{Step 1: Analyze the street-view photo.} The photo is taken from the center of an intersection, looking straight ahead. On the immediate right is a very tall, light-colored (white or beige) apartment or office building. \\
      It runs along the street from the corner onwards. On the immediate left is a fenced-off open area, which looks like a park or a sports field. \\
      There are trees lining the sidewalk. \\
      In the distance, further down the road, at least two prominent skyscrapers are visible. The road itself is a multi-lane street. \\
      \textbf{Step 2: Analyze the overhead map and evaluate each option.} The camera is at the central intersection. \\
      Option A (looking East): To the right (south) is a large white building. To the left (north) is another large white building. This does not match the photo, which has a park/field on the left. \\
      Option B (looking South): To the right (west) is a large white building. To the left (east) is another large white building. This does not match the photo. \\
      Option C (looking West): To the right (north) is a large green area with a baseball field. To the left (south) is a large white building. This is the opposite of the street view (building on the right, park on the left). \\
      Option D (looking North): To the right (east) is a long, rectangular, light-colored building located right at the corner. To the left (west) is a large green area containing a baseball field. This arrangement perfectly matches the street-view photo. \\
      \textbf{Step 3: Conclude the comparison.} Direction D is the only view from the intersection that has a large park/baseball field on the left and a tall, long building on the right. Final answer: \textbackslash boxed\{D\}. \\
      \midrule
      \textbf{Ground truth} \\
      D \\
      \bottomrule
    \end{tabular}
  \end{minipage}

  \caption{Illustrative m2sv example with paired images and a model-generated
  reasoning trace. The model correctly infers the viewing direction by aligning
  stable map geometry with the Street View image.}
  \label{fig:m2sv-example}
\end{figure*}

\section{Introduction}

Vision--language models have demonstrated impressive capabilities in object
recognition, diagram understanding, and text-heavy multimodal reasoning.
However, they continue to struggle with tasks that require consistent spatial
grounding across viewpoints. In particular, aligning abstract overhead
representations such as maps with egocentric, ground-level imagery remains a
fundamental challenge. This capability is essential for applications in
navigation, robotics, and embodied assistants, yet is weakly represented in
current evaluation benchmarks.

Most existing multimodal reasoning benchmarks emphasize synthetic diagrams,
charts, or carefully curated visual puzzles where geometry is simplified or
explicitly annotated. Real-world spatial reasoning is substantially harder: it
requires coping with viewpoint changes, partial observability, temporal mismatch
between data sources, and ambiguity arising from repeated or symmetric structures.
As a result, current benchmarks may overestimate spatial competence by avoiding
the alignment of disparate viewpoints. We reserve a deeper discussion of related
work and background for Appendix~\ref{sec:background}.

In this work, we introduce \textbf{m2sv} (map-to-street-view), a benchmark designed
to isolate a core spatial reasoning primitive: inferring camera viewing direction
by aligning a north-up overhead map with a Street View image captured at the same
real-world intersection (Figure~\ref{fig:m2sv-example}). Each example requires
models to reason about road topology, intersection geometry, and stable landmarks,
while ignoring transient or unreliable cues such as vehicles, lighting, or
pedestrians. By grounding questions in real-world imagery and geometry, m2sv
exposes failure modes that are largely hidden in synthetic or single-view tasks.

Beyond serving as a benchmark, m2sv enables a systematic study of \emph{difficulty}
in spatial reasoning. Because intersection geometry and candidate directions are
known a priori, we can define scalable structural difficulty signals such as the
number of candidate directions and angular symmetry. We complement these with
human response-time measurements, allowing us to distinguish between structural
ambiguity and human-perceived difficulty. This dual view reveals that even
moderate geometric ambiguity induces sharp performance drops in models, while
humans remain near ceiling.

Finally, we perform an extensive failure analysis of our best adapted open model,
going beyond aggregate accuracy to identify recurring
breakdowns such as egocentric--allocentric confusion, symmetry traps, unreliable
cue reliance, and internal reasoning inconsistencies.

Our contributions are:
\begin{itemize}
  \item A fully automated pipeline for constructing large-scale map-to-street-view
  spatial reasoning benchmarks, including image data sourcing, candidate response
  generation, and labeling tools.
  \item \textbf{m2sv-20k}, a geographically diverse benchmark with controlled
  ambiguity and statistically reliable evaluation, and \textbf{m2sv-sft-11k}, a
  curated set of structured reasoning traces for supervised fine-tuning.
  \item A systematic characterization of \textbf{difficulty in spatial reasoning},
  combining structural signals (e.g., number of candidate directions and angular
  symmetry) with human effort and model behavior (Section~\ref{sec:analysis-difficulty}).
  \item An extensive evaluation and \textbf{failure mode analysis}, identifying
  recurring breakdowns such as egocentric--allocentric confusion, symmetry traps,
  unreliable cue reliance, and internal reasoning inconsistencies
  (Section~\ref{sec:failure-analysis}).
\end{itemize}

\paragraph{Reproducibility and open release.}
All artifacts are publicly available at our project repository.\footnote{Code,
data, and interface: \url{https://github.com/yosubshin/m2sv}. Live
human-evaluation interface: \url{https://m2sv.yosubshin.com}.} We release: (i) the
\textbf{m2sv-20k} blueprints---intersection coordinates, candidate azimuths,
ground-truth labels, and Street View panorama identifiers---together with the
deterministic rendering pipeline, so the full benchmark can be reconstructed on
demand under licensing constraints; (ii) the \textbf{m2sv-sft-11k} reasoning
traces; (iii) all evaluation and fine-tuning code, including prompts and
hyperparameters; and (iv) the interactive human-evaluation interface, hosted live
at \url{https://m2sv.yosubshin.com}, along with the collected per-item human
labels and response times. Together these enable reproduction of the zero-shot,
adaptation, and human-baseline results reported in this paper.

\section{Benchmark Design}

Each m2sv example consists of two images: (1) a north-up overhead map centered at
a real-world intersection, annotated with labeled directional rays, and (2) a
Street View image captured near the same intersection and oriented along one of
those rays. The task is to identify which labeled direction corresponds to the
Street View camera orientation.

The number of candidate directions per example ranges from 2 to 7, with a median
of 3 (mean: 3.24). Accordingly, the random-guess baseline is approximately
31.4\%. Accuracy is measured over multiple-choice labels.

\section{Dataset and Blueprint Pipeline}

\subsection{Blueprint specification}

We decouple dataset specification from image rendering by releasing blueprint
metadata that records intersection coordinates, candidate azimuths, correct
labels, and Street View panorama identifiers. Images are rendered on demand using
a standardized pipeline, enabling deterministic reconstruction of the dataset.

\subsection{Sampling and filtering}

Intersections are sampled from a seed list of major metropolitan areas worldwide
to encourage broad geographic coverage. The sampling pipeline itself is
city-agnostic and can be readily extended to additional regions by modifying the
city list.

To control ambiguity and ensure fair evaluation, we apply several filtering
criteria. Street View panoramas are restricted to within 5 m of the intersection
center, resulting in a tightly controlled offset distribution (median: 2.05 m;
95th percentile: 4.51 m). We further enforce a minimum angular separation of 15°
between candidate azimuths to exclude near-collinear road configurations that are
difficult even for humans to disambiguate.

\subsection{Dataset statistics}

m2sv-20k contains 20,000 examples (10k train, 10k validation) spanning 32 cities across multiple continents.
Per-city caps prevent over-representation of dense urban areas. The number of
candidate directions per example ranges from 2 to 7 (min: 2, max: 7, mean: 3.24,
median: 3). Table~\ref{tab:dataset-stats} in the appendix reports additional statistics.

\subsection{Reasoning traces}

We curate \textbf{m2sv-sft-11k}, a subset of 11,000 examples annotated with
structured reasoning traces generated by Gemini-2.5-Pro. Traces explicitly
compare map geometry and Street View imagery and terminate in a boxed final
answer. Fine-tuning on all traces slightly underperformed, so we filter to
correct traces only, yielding 4.4k examples total (4.37k train, 485 validation).
These traces are used exclusively for supervised fine-tuning experiments.

\section{Experimental Setup}

We evaluate a range of proprietary and open-source VLMs using a standardized
prompt that describes the task and instructs models to ignore transient visual
elements. Accuracy and 95\% confidence intervals are reported.

We fine-tune from Qwen3-VL-8B-Instruct using LoRA for both SFT and RL. SFT uses
the filtered correct-trace subset (4.4k) and LoRA improves accuracy by 1.2\%
over full-parameter SFT. We train with LR 2e-4, cosine schedule with 0.03 warmup
ratio, batch size 4, and early stopping at 2.3 epochs (max 4), using 1xH200 for
about 2 hours. RL starts from the SFT checkpoint and uses LoRA with LR 2e-5,
generation\_batch\_size 64, per\_device\_train\_batch\_size 16, and gradient\_accumulation\_steps 16,
trained on 2xH200 for about 36 hours; we report results from the step-550
checkpoint.


\section{Results}
\label{sec:results}

\subsection{Zero-shot baselines and the human performance gap}
\label{sec:results-zeroshot}

We begin by establishing zero-shot baselines and the human performance ceiling on
m2sv. Table~\ref{tab:m2sv-baselines} reports accuracy for a range of proprietary
and open vision--language models evaluated without task-specific adaptation,
alongside a random baseline and human performance measured on a 200-example
subset.

\begin{table}[t]
  \centering
  \caption{Zero-shot baseline performance on m2sv with 95\% confidence intervals.
  The engaged-human row reports the mean over eight annotators with the
  between-annotator standard deviation (not a binomial interval).}
  \label{tab:m2sv-baselines}
  \begin{tabular}{l r r}
    \toprule
    Model & N & Accuracy (\%) \\
    \midrule
    Gemini-3-Pro & 1k & 65.2 $\pm$ 3.0 \\
    GPT-5 & 1k & 57.2 $\pm$ 3.1 \\
    Gemini-2.5-Pro & 1k & 47.2 $\pm$ 3.1 \\
    \midrule
    Qwen3-VL-8B-Instruct & 1k & 35.5 $\pm$ 3.0 \\
    Qwen3-VL-235B-A22B-Thinking & 1k & 42.7 $\pm$ 3.1 \\
    \midrule
    Random baseline & 1k & 31.4 $\pm$ 2.9 \\
    \midrule
    Human (engaged, $n{=}8$) & 200 & 72.0 $\pm$ 9.7 \\
    Human (expert) & 200 & 95.0 \\
    \bottomrule
  \end{tabular}
\end{table}

To establish a robust human reference, we conduct an independent multi-annotator
study on the 200-example subset. Twelve annotators---none of whom participated in
dataset construction or had prior exposure to the benchmark
items\footnote{Eleven are external volunteers; the twelfth is a co-author who
contributed to the reinforcement-learning pipeline but not to the benchmark data
or the human-evaluation protocol, and encountered the items for the first time
during the study.}---answer items through a web interface that logs the selected
option and per-item response time. Ten completed all 200 items and form our
analysis set; the remaining two answered only 17 and 76 items and are excluded for
comparability (their small overlap also makes pairwise agreement unstable). We
additionally retain the original expert annotator (a dataset author) as a ceiling
reference.

\begin{figure}[t]
  \centering
  \includegraphics[width=\linewidth]{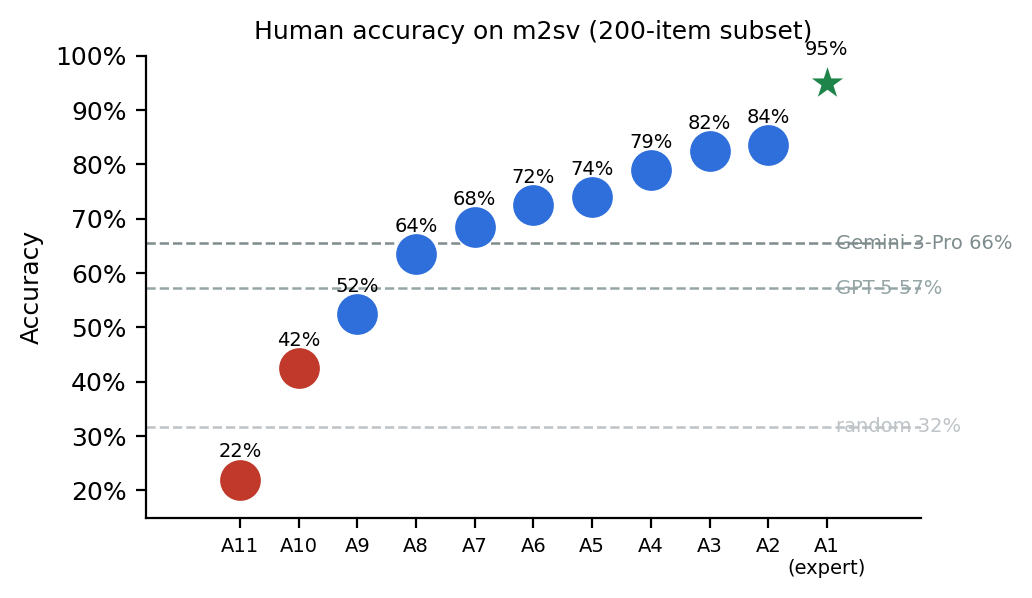}
  \caption{
    \textbf{Per-annotator human accuracy on the 200-example subset}
    (the ten annotators who completed all 200 items). Annotators span
    $22\%$--$84\%$; the two low-agreement outliers ($\kappa<0.3$ with the expert)
    are shown in red and the expert as a star. Dashed lines denote model and random
    baselines.
  }
  \label{fig:human-dist}
\end{figure}

\begin{figure}[t]
  \centering
  \includegraphics[width=\linewidth]{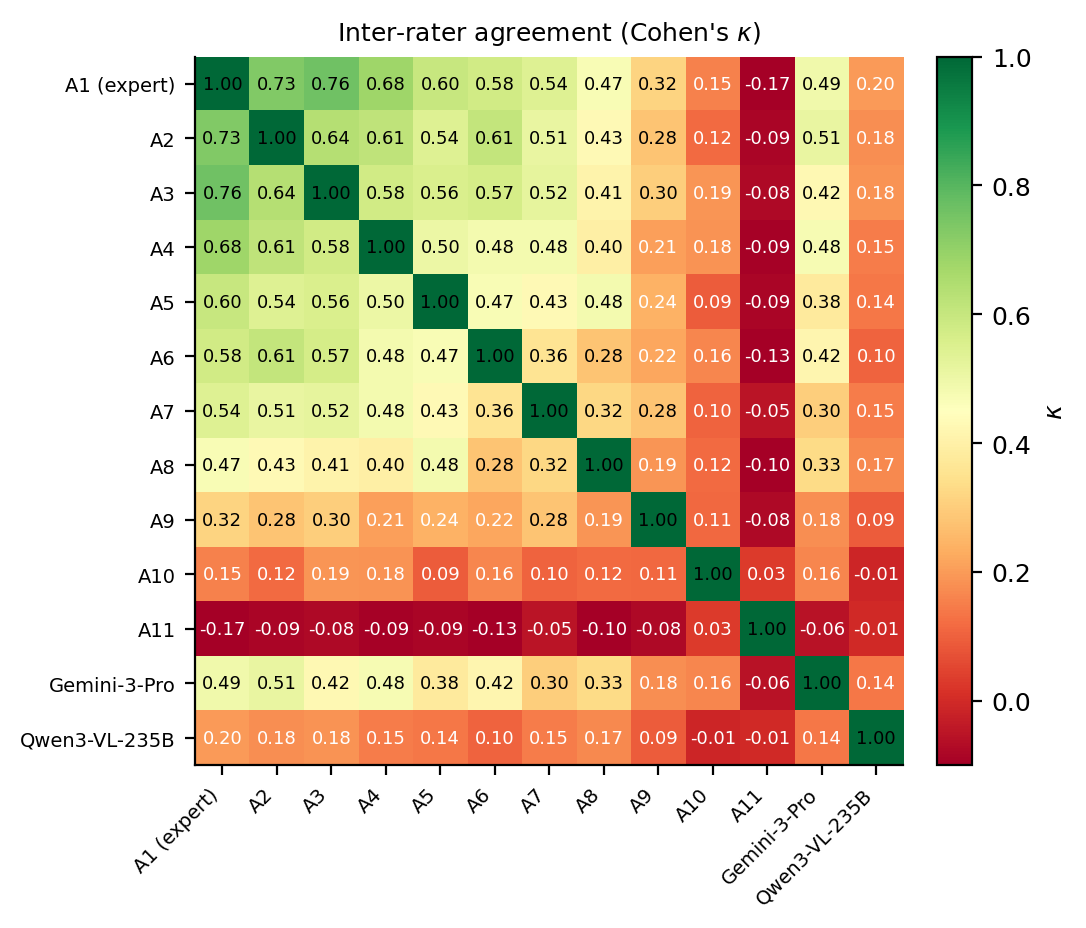}
  \caption{
    \textbf{Inter-rater agreement (Cohen's $\kappa$),} chance-corrected per item
    by candidate count. The expert (A1) and engaged annotators (A2--A9) form a
    high-agreement block; Gemini-3-Pro sits below the human consensus and
    Qwen3-VL-235B lower still. The low-agreement annotators (A10, A11) are outliers against
    all raters.
  }
  \label{fig:kappa}
\end{figure}

Human accuracy is high but effort-dependent
(Figure~\ref{fig:human-dist}). Eight mutually consistent annotators average
$72.0\%$ (between-annotator SD $9.7\%$) and the expert reaches $95\%$, all above
the strongest model (Gemini-3-Pro, $65.2\%$); two annotators answered at or below
chance ($42\%$ and $22\%$) and are flagged not by accuracy but by near-zero or
negative agreement with every other rater (Cohen's $\kappa \le 0.15$)---one was
fast and one slow, so the exclusion reflects disagreement, not speed. The
below-chance annotator's answers were systematically the \emph{opposite} heading
(median angular error $150^\circ$, negative $\kappa$); a debrief traced this to a
misunderstood convention---treating the labeled marker as the camera's location
rather than the intersection center, which inverts every heading---rather than a
spatial-reasoning deficit, plausibly compounded by non-native instructions.
Agreement-based exclusion thus removes task misunderstanding that an accuracy
threshold alone would conflate with poor spatial ability.
Inter-annotator
agreement is otherwise moderate to substantial (Figure~\ref{fig:kappa}): the
expert and the engaged annotators agree well with one another (median pairwise
$\kappa = 0.48$, up to $0.76$), confirming that the labels are reliable and that
attentive humans converge on the same answers. In contrast, Gemini-3-Pro agrees
with humans only moderately (median $\kappa = 0.42$, below the human--human
median) and Qwen3-VL-235B shows only slight agreement (median $\kappa = 0.15$). The human--model gap is
thus visible not only in accuracy but in agreement structure: the best model sits
below the human consensus, indicating that zero-shot pretrained VLMs
lack robust mechanisms for aligning overhead maps with egocentric street-level
views.

\subsection{Adaptation baselines on m2sv}
\label{sec:results-adaptation}

We next evaluate whether task-specific adaptation can mitigate this gap.
Table~\ref{tab:m2sv-main-results} reports results for Qwen3-VL-8B under three
training regimes: the base model, supervised fine-tuning (SFT) on m2sv reasoning
traces, and reinforcement learning (RL) applied on top of SFT.

\begin{table}[t]
  \centering
  \caption{Adaptation results on m2sv-20k (10k validation).}
  \label{tab:m2sv-main-results}
  \begin{tabular}{l l r}
    \toprule
    Model & Training & Accuracy (\%) \\
    \midrule
    Qwen3-VL-8B & Base & 34.3 $\pm$ 0.9 \\
    Qwen3-VL-8B & SFT & 39.8 $\pm$ 1.0 \\
    Qwen3-VL-8B & SFT+RL & 43.9 $\pm$ 1.0 \\
    \bottomrule
  \end{tabular}
\end{table}

Supervised fine-tuning yields a consistent improvement over the base model, and
reinforcement learning provides an additional gain of approximately four points.
These results demonstrate that m2sv-specific supervision teaches useful task
structure and improves decision consistency. However, even after SFT and RL,
performance remains more than 50 points below the human ceiling, indicating that
adaptation alone does not close the gap in spatial alignment capability.

\subsection{Cross-benchmark transfer}
\label{sec:results-transfer}

To assess whether improvements on m2sv reflect more general spatial reasoning
abilities, we evaluate m2sv-trained models on established multimodal benchmarks.
Table~\ref{tab:transfer} in the appendix reports performance on a subset of commonly used
benchmarks for Qwen3-VL-8B variants.

Transfer results are mixed and do not show consistent improvements across
benchmarks. Some metrics improve under SFT or SFT+RL, while others degrade or
remain unchanged. This sensitivity suggests that gains on m2sv do not reliably
translate into broader multimodal reasoning improvements, and may instead reflect
benchmark-specific adaptations.

\subsection{Asymmetric transfer with MindCube}
\label{sec:results-mindcube}

To further probe transferability, we conduct a focused evaluation on the MindCube
suite~\citep{yin2025spatialmental}, which also targets spatial reasoning but
differs in input structure and output format. Table~\ref{tab:m2sv-generalization} in the appendix reports performance under
multiple prompting configurations for Qwen3-VL-8B variants.

m2sv-trained models occasionally outperform the base model on MindCube, but gains
are highly sensitive to prompting format and are largest in \emph{direct answer}
settings. When explicit reasoning traces are requested, the base model matches or
exceeds the adapted variants. Conversely, models trained on MindCube perform near
random on m2sv, indicating asymmetric and
limited transfer.

Together, these results suggest that current fine-tuning pipelines induce
benchmark-specific reasoning policies rather than transferable spatial
representations, despite surface-level similarity between tasks.

\section{Measuring Difficulty in Map-to-Street-View Reasoning}
\label{sec:analysis-difficulty}

A central challenge in evaluating spatial reasoning is defining what makes an
example \emph{difficult}. In the map-to-street-view task, difficulty does not arise
from a single factor but from an interaction between intersection geometry and
fine-grained visual confusability across candidate directions. We therefore
distinguish between two complementary notions of difficulty:
\emph{structural difficulty}, which is automatically measurable and scalable, and
\emph{human-perceived difficulty}, which reflects the cognitive effort required to
resolve subtle spatial ambiguities. We reflect on the background of vision-language spatial reasoning difficulty analysis in Appendix~\ref{sec:prior-difficulty}.

Structural measures are attractive for large-scale dataset analysis and potential
curriculum learning, but they are necessarily coarse. Human-perceived difficulty,
while expensive to obtain, provides a diagnostic signal that captures visual and
semantic factors beyond geometry. We analyze both to characterize where current
vision--language models succeed and where they fail.

\subsection{Structural difficulty signals}
\label{sec:analysis-difficulty-structural-signals}

\subsubsection{Structural Difficulty via Number of Candidate Directions}
\label{sec:analysis-options}

We analyze difficulty through a structural lens using the number of candidate
directions (\#options) provided in each question. This quantity is derived
directly from intersection geometry and is available for every example without
human annotation, making it a scalable proxy for geometric ambiguity and a
natural signal for dataset filtering or curriculum learning.

A direct comparison of raw accuracy across different values of \#options is
confounded by varying chance levels (e.g., $0.5$ for two options vs.\ $0.25$ for
four options). To account for this, we report \emph{chance-normalized accuracy
gain}, defined as $(a - c)/(1 - c)$, where $a$ is model accuracy and
$c = 1/K$ is the chance accuracy for $K$ options. This normalization measures how
much useful signal a model extracts beyond random guessing.

\begin{figure}[t]
  \centering
  \includegraphics[width=\linewidth]{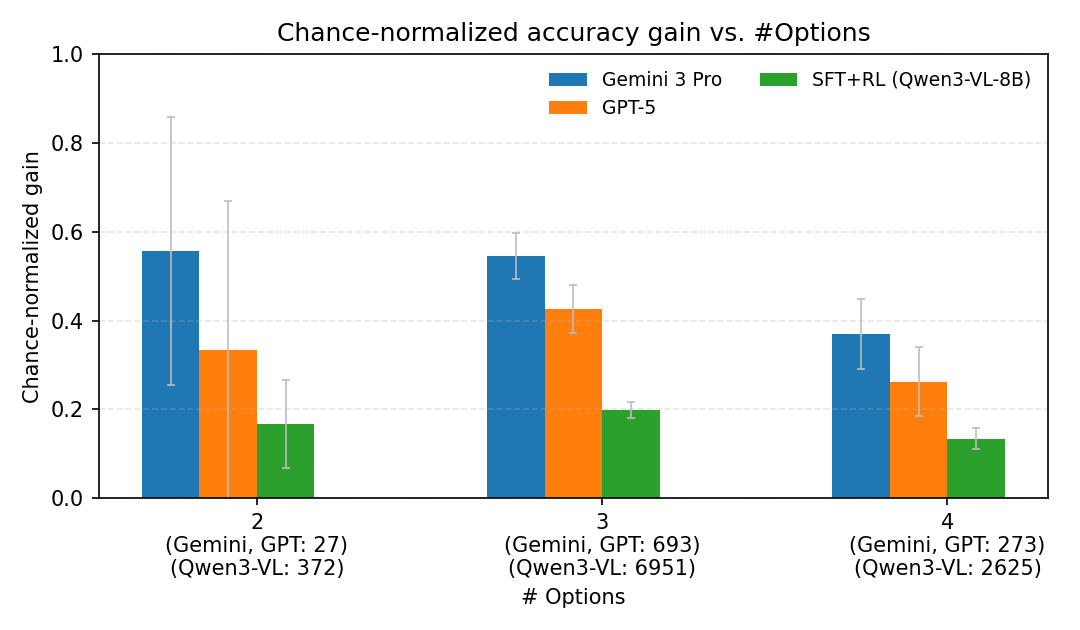}
  \caption{
    \textbf{Chance-normalized accuracy vs.\ structural difficulty.}
    Accuracy normalized by chance level as a function of the number of candidate
    directions (\#options).
    Qwen3-VL-8B (SFT+RL) is evaluated on the full 10k validation set, while
    Gemini~3~Pro and GPT-5 are evaluated on a 1k subset.
    Three-option intersections yield the highest normalized gains across models,
    reflecting the benefit of geometric asymmetry.
    Error bars denote 95\% binomial confidence intervals.
  }
  \label{fig:accuracy-vs-options}
\end{figure}

Figure~\ref{fig:accuracy-vs-options} reports chance-normalized performance as a
function of the number of options (2, 3, and 4) for our best open model
(Qwen3-VL-8B SFT+RL) evaluated on the full 10k validation set, and for two strong
proprietary models evaluated on a 1k subset. This enables a statistically stable,
large-scale analysis of structural difficulty.

Across all models, normalized performance does \emph{not} degrade monotonically
with the number of options. Instead, three-option intersections yield the highest
normalized gains, outperforming both two- and four-option cases. This pattern is
consistent across open and proprietary systems and reflects the role of
geometric asymmetry: three-way intersections are often T-junctions with
distinctive spatial layouts, whereas two- and four-way intersections are more
likely to be symmetric and thus ambiguous.

These results demonstrate that the number of candidate directions alone is an
incomplete measure of structural difficulty. Intersection cardinality interacts
with geometric symmetry, and higher difficulty arises not simply from having
more choices, but from the absence of asymmetry that can break directional
ambiguity. This motivates the finer-grained structural and perceptual difficulty
analyses that follow.

\subsubsection{Structural Ambiguity via Road-Azimuth Symmetry}
\label{sec:analysis-symmetry}

The number of options captures how many candidate directions are present, but not
how \emph{interchangeable} those directions are. To refine the notion of
structural difficulty, we introduce a road-azimuth symmetry measure based on the
angular spacing between candidate directions. Intersections with evenly spaced
azimuths (e.g., four-way junctions) are geometrically symmetric, while those with
highly uneven spacing exhibit a dominant road direction and are structurally
asymmetric.

Concretely, let $\theta_1 \le \cdots \le \theta_K$ be the $K$ candidate azimuths
(in degrees) sorted on the circle, and define the $K$ consecutive angular gaps
\begin{equation}
  g_i = \theta_{i+1} - \theta_i \;\; (1 \le i < K), \qquad
  g_K = 360^\circ + \theta_1 - \theta_K ,
\end{equation}
where $g_K$ is the wrap-around gap. We define the \emph{road-azimuth symmetry} of
an intersection as the ratio of its largest to smallest gap,
\begin{equation}
  S \;=\; \frac{\max_i g_i}{\min_i g_i} \;\ge\; 1 .
  \label{eq:symmetry}
\end{equation}
A regular intersection with evenly spaced rays has $S = 1$ (maximally symmetric),
whereas a dominant-road layout containing one large gap has $S \gg 1$
(asymmetric). Two properties of $S$ require care. First, because candidate
azimuths are coarsely quantized, $S$ takes few distinct values---a single value
$S=2$ accounts for nearly half of all three-option intersections---so fine-grained
quantile binning arbitrarily splits ties. Second, symmetry is correlated with
candidate count: the most symmetric intersections ($S\approx 1$) are
overwhelmingly four-way junctions. To avoid both pitfalls, we hold candidate count
fixed---analyzing the three-option stratum, the largest group---and bin into three
well-separated, untied levels: \emph{Symmetric} ($S<2$), \emph{Intermediate}
($S=2$), and \emph{Asymmetric} ($S>2$).

\begin{figure}[t]
  \centering
  \includegraphics[width=\linewidth]{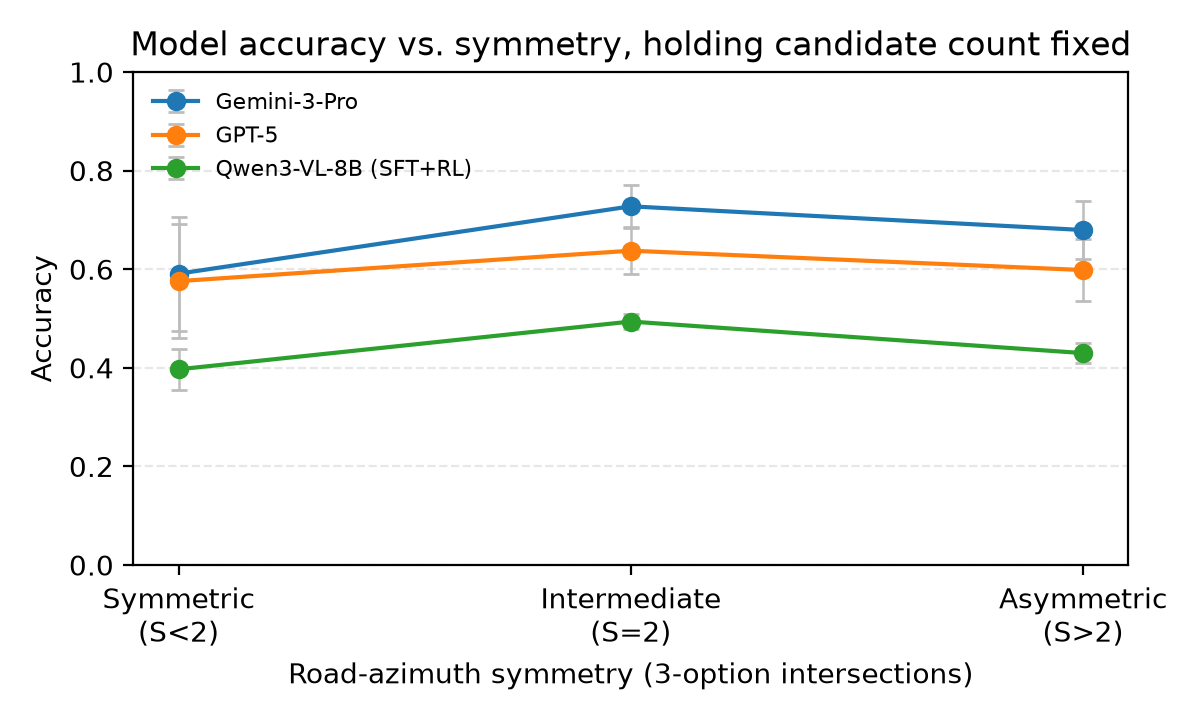}
  \caption{
    \textbf{Model accuracy vs.\ road-azimuth symmetry (three-option
    intersections).}
    Accuracy by symmetry level $S$ (Eq.~\ref{eq:symmetry}) with candidate count
    held fixed, removing the confound between symmetry and the number of options.
    All three models are hardest on symmetric ($S<2$) intersections and peak at
    intermediate symmetry ($S=2$).
    Error bars denote 95\% binomial confidence intervals.
  }
  \label{fig:accuracy-vs-symmetry}
\end{figure}

Figure~\ref{fig:accuracy-vs-symmetry} reports model accuracy across these bins.
Even with candidate count held fixed, accuracy is non-monotonic in symmetry: all
three models are hardest on the most \emph{symmetric} intersections (e.g.,
Gemini-3-Pro $59\%$ vs.\ $73\%$ at intermediate symmetry), with a smaller decline
toward the most asymmetric layouts. The dip at high symmetry is consistent across
open and proprietary systems, confirming that near-even (Y-shaped) intersections
are genuinely the most confusable rather than an artifact of candidate count.

This pattern indicates that azimuth symmetry is a real but non-monotonic driver of
difficulty. Near-even intersections offer no dominant road to anchor the candidate
directions and are the hardest; a single straight through-road (intermediate
symmetry) is the most recognizable; and strongly dominant-road layouts are
slightly harder again. Symmetry thus bounds difficulty without fully determining
it.

\subsection{Human effort as a difficulty proxy}
\label{sec:analysis-human-time}

Structural measures do not capture visual confusability between candidate
directions, which often determines whether fine-grained spatial alignment is
required. To probe these effects directly, we analyze difficulty as perceived by
human annotators, using response time as a proxy for cognitive effort.

Using the multi-annotator study of Section~\ref{sec:results-zeroshot}, we record
per-item response time for every annotator on the 200-example subset. Unlike
models, humans show no symmetry-dependent slowdown: per-problem median response
time is essentially flat across symmetry ($\sim$15\,s in every three-option
symmetry bin), so humans resolve symmetric and asymmetric intersections with
comparable effort even though models collapse on the symmetric ones
(Figure~\ref{fig:accuracy-vs-symmetry}).

\begin{figure}[t]
  \centering
  \includegraphics[width=\linewidth]{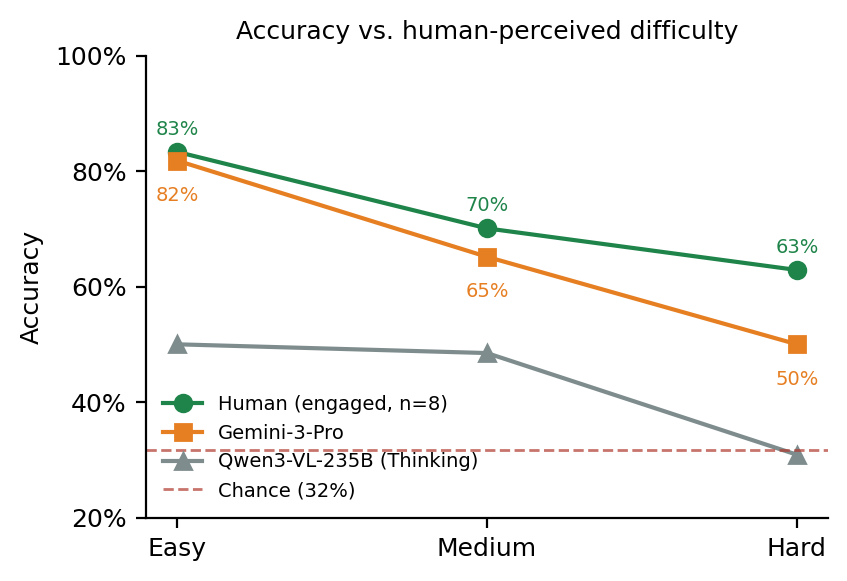}
  \caption{
    \textbf{Accuracy vs.\ human-perceived difficulty (multi-annotator).}
    Examples are partitioned into Easy/Medium/Hard tertiles by \emph{median}
    response time across the engaged annotators who completed all 200 items. The
    difficulty axis is a human-effort signal independent of any model, so the model
    curves and the widening gap are not artifacts of the binning; the human curve is
    shown for reference. Humans degrade gently and stay above both models on every
    tier; both models lose the most ground on the hardest tier---Gemini-3-Pro falls
    toward, and Qwen3-VL-235B to, chance---where the human--model gap is largest.
  }
  \label{fig:accuracy-vs-human-time}
\end{figure}

Figure~\ref{fig:accuracy-vs-human-time} instead uses response time as a global
difficulty signal: examples are partitioned into tertiles by the \emph{median}
response time across the engaged annotators who completed all 200 items (the
median is robust to occasional idle responses, which the visibility detector
cannot catch). This difficulty axis is a purely human-effort signal, computed
without reference to any model, so the model accuracy curves---and hence the
human--model gap---cannot be artifacts of how difficulty is defined. Taking the
median across many annotators, rather than relying on a single rater's latency as
in our pilot, also removes the within-subject coupling between one rater's
per-trial speed and correctness that would otherwise be conflated with item
difficulty. The human accuracy curve is shown for reference: that effort and
accuracy co-vary within the same annotators is expected and is not the claim we
draw from this figure. Humans degrade gently with difficulty ($83\%$ to $63\%$)
and remain well above both models on every tier; the models, by contrast, lose the
most ground precisely on the hardest examples. Gemini-3-Pro degrades more steeply
than humans at \emph{both} steps ($82\%\!\to\!65\%\!\to\!50\%$), so the human--model
gap widens monotonically---Gemini nearly matches humans on the easiest tier but
trails by $13$ points on the hardest. Qwen3-VL-235B (Thinking) instead holds
roughly flat through the easy and medium tiers ($50\%$, $48\%$) and then collapses
to chance on the hardest tier ($31\%$ vs.\ $\sim$30\%); its degradation is
concentrated entirely in the hard regime, where humans still reach $63\%$. The
common pattern is thus a hardest-tier collapse: both models fall to (Qwen) or
toward (Gemini) chance exactly where fine-grained spatial alignment is most
required, while humans hold up. Candidate-count composition is similar across
bins, so chance is $\sim$30--32\% in all three tiers.

Taken together, these results suggest that model failures are driven not by
geometry alone, but by an inability to exploit subtle, stable cues when visual
ambiguity remains high. While structural asymmetry allows humans to shortcut the
alignment process, current vision--language models do not reliably adapt their
inference strategy to difficulty, motivating future work on difficulty-aware
training and adaptive inference.

\subsection{Model behavior vs. difficulty}
\label{sec:analysis-model-behavior-vs-difficulty}

We analyze how models modulate the length of their reasoning traces as a function
of problem difficulty. As the difficulty axis we reuse the same multi-annotator
median-response-time tertiles as Figure~\ref{fig:accuracy-vs-human-time}.
Trace length is measured as the approximate number of generated tokens in the
model’s reasoning output.

\begin{figure}[t]
  \centering
  \includegraphics[width=\linewidth]{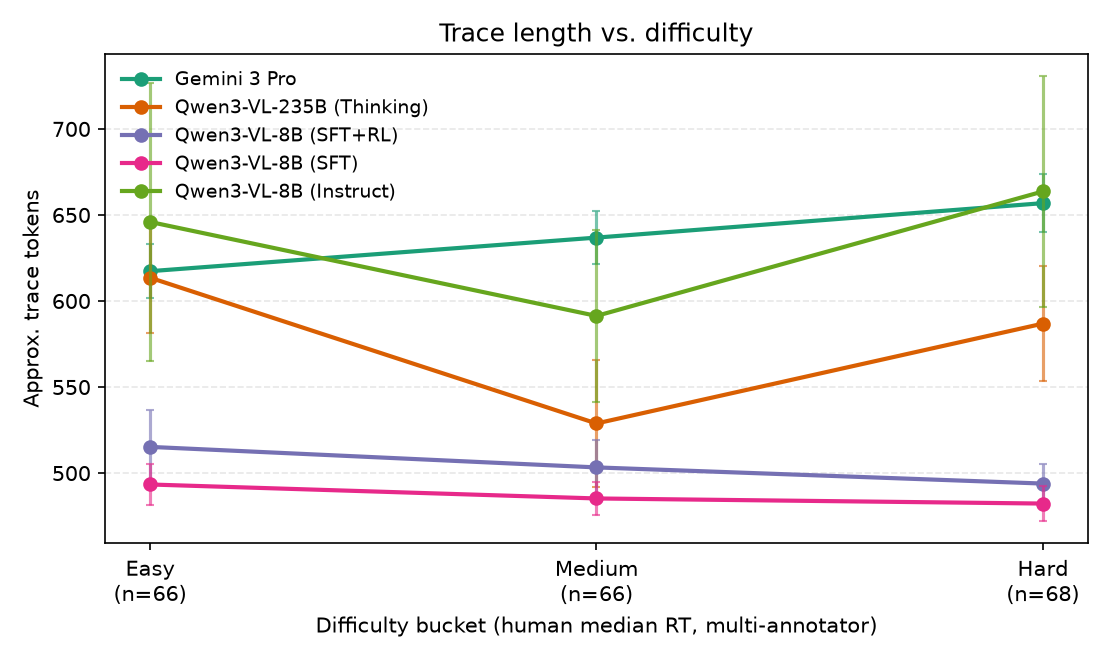}
  \caption{
    \textbf{Reasoning trace length vs.\ difficulty.}
    Average trace length (in tokens) as a function of difficulty (multi-annotator
    median-response-time tertiles, the same bins as
    Figure~\ref{fig:accuracy-vs-human-time}).
    Gemini~3~Pro and Qwen3-VL-235B (Thinking) increase trace length on harder
    examples, while Qwen3-VL-8B models adapted via SFT and SFT+RL produce
    near-constant-length traces despite higher accuracy.
    Error bars denote standard error of the mean.
  }
  \label{fig:trace-length-vs-difficulty}
\end{figure}

Figure~\ref{fig:trace-length-vs-difficulty} reports average trace length across
difficulty buckets (easy, medium, hard) for proprietary models and multiple
variants of Qwen3-VL.
We exclude GPT-5 from this analysis, as we observed that it does not reliably
produce explicit reasoning traces and often responds with a direct answer only,
making trace-length measurements inconsistent.

Gemini~3~Pro expands its reasoning on harder examples, increasing trace length
monotonically across difficulty; Qwen3-VL-235B (Thinking) trends upward overall
but less cleanly. The base Qwen3-VL-8B Instruct model produces long traces but
without a consistent difficulty trend.

In contrast, both the supervised fine-tuned Qwen3-VL-8B (SFT) and the
Qwen3-VL-8B (SFT+RL) models produce \emph{markedly shorter and nearly
constant-length} traces across difficulty levels. Adaptation thus compresses
reasoning toward a short, fixed budget rather than preserving the
difficulty-adaptive expansion exhibited by the strongest model, even though
overall accuracy improves.

Taken together, trace length modulation and task accuracy appear to be separable
capabilities.
Stronger proprietary models retain both difficulty awareness and effective
multi-cue integration, while adapted open models achieve higher accuracy at the
cost of reduced adaptive reasoning depth.
This behavior aligns with the failure modes identified in
Section~\ref{sec:failure-analysis}, where adapted models often commit prematurely
under symmetry or rely on a single dominant feature instead of escalating to
finer-grained spatial evidence.

\section{Failure Mode Analysis}
\label{sec:failure-analysis}

To better understand why current vision--language models fail on m2sv despite
strong zero-shot performance on other multimodal benchmarks, we conduct a
qualitative failure analysis on a subset of challenging examples. Rather than
attempting an exhaustive taxonomy, we focus on a small number of recurring
failure patterns that consistently appear across models and difficulty regimes.

Importantly, these failures do not reflect fundamentally unsolvable cases: human
annotators are able to solve nearly all examples, often by investing additional
effort and exploiting subtle but stable spatial cues. Model errors instead arise
from brittle alignment strategies and inappropriate reliance on unreliable visual
evidence.

Below, we summarize common failure modes observed in both open and
proprietary models, illustrated with representative examples.

\paragraph{Egocentric--allocentric confusion (left--right inversion).}
Models frequently misinterpret the relationship between egocentric directions in
the street-view image and allocentric directions on the overhead map. This often
manifests as left--right inversions, where a correct identification of landmarks
is followed by an incorrect mapping to map orientation, particularly at
four-way intersections (Fig.~\ref{fig:failure-examples}a).

\paragraph{Over-reliance on unreliable cues.}
In many failures, models base eliminations on visually salient but unstable cues
such as roof color, shadows, lighting conditions, or image blur. These cues are
often weakly or not at all represented in overhead imagery and can vary
substantially with capture time, leading to confident but incorrect conclusions
(Fig.~\ref{fig:failure-examples}b).

\paragraph{Landmark misbinding.}
Models sometimes correctly identify salient landmarks in the overhead view but
misassociate them with the wrong candidate direction or parcel. For example, a
swimming pool visible several houses away may be incorrectly attributed to the
immediate intersection, leading to erroneous eliminations. This suggests limited
precision in spatial binding between landmarks and their local context
(Fig.~\ref{fig:failure-examples}c).

\paragraph{False landmark projection.}
In other cases, models introduce landmarks that are not present within the local
intersection context, such as distant structures or perspective-induced artifacts.
These projected landmarks are then treated as decisive evidence, despite lacking
support in the overhead map, indicating weak control over spatial scope and
relevance (Fig.~\ref{fig:failure-examples-appendix}d).

\paragraph{Failure to maintain a consistent spatial representation.}
In some failures, models do not maintain globally consistent spatial properties
across options, treating shared structures as independent entities. This can
manifest as assigning different road widths, lane counts, or landmark layouts to
opposite headings of the same road, or contradicting earlier correct spatial
descriptions during later reasoning. These errors suggest that models evaluate
options in isolation rather than constructing and enforcing a coherent global
spatial model (Fig.~\ref{fig:failure-examples-appendix}a).

\paragraph{Internal contradiction and state inconsistency.}
In some failures, models correctly identify spatial relationships or landmark
configurations, but subsequently contradict themselves while reasoning or
verbalizing the solution. These errors are characterized by internally
inconsistent statements (e.g., asserting a correct left--right relation and then
rejecting it), even though the underlying spatial alignment is correct. Unlike
egocentric--allocentric confusion, these failures arise from an inability to
maintain and consistently apply inferred spatial commitments across reasoning
steps (Fig.~\ref{fig:failure-examples-appendix}b).

\paragraph{Symmetry traps and insufficient cue escalation.}
At symmetric or near-symmetric intersections, models often identify multiple
plausible candidate directions but fail to escalate to finer-grained geometric
cues—such as lane counts, road markings, curb geometry, or persistent vegetation—
that would resolve the ambiguity. Instead, models rely on weak or unstable
evidence and prematurely eliminate the correct option
(Fig.~\ref{fig:failure-examples-appendix}c).

\subsection{Implications.}
Taken together, our analyses suggest that model failures on m2sv are not driven
primarily by geometric complexity or lack of visual information, but by an
inability to adapt inference strategies to the level of ambiguity present in an
example. Structural asymmetry allows humans to shortcut reasoning on easy cases,
while symmetric or visually confusable cases trigger deeper, more careful
comparison. Current vision--language models, by contrast, exhibit relatively
uniform reasoning behavior across difficulty regimes, leading to brittle
decisions when subtle spatial cues are required.

This gap highlights an important limitation of current training and inference
pipelines: models do not reliably escalate their reasoning depth or evidence
aggregation in response to increased difficulty. Addressing this limitation may
require difficulty-aware training curricula, adaptive inference mechanisms, or
explicit modeling of uncertainty in spatial alignment tasks.

\section{Limitations and Future Work}
\label{sec:limitations}

\textbf{m2sv} relies on the availability and coverage of commercial mapping services
and consequently inherits biases from their data collection practices, including
uneven geographic coverage and representation. The benchmark is restricted to
public road intersections and does not include private, restricted, or sensitive
locations.

Our human evaluation comprises ten annotators together with one expert on
the 200-example analysis subset, with inter-annotator agreement reported in
Section~\ref{sec:results-zeroshot}. While this is sufficient to characterize
between-annotator variance and to de-confound the response-time difficulty
analysis, the annotator pool remains modest and was recruited opportunistically
rather than through a controlled, pre-registered protocol; a larger and more
demographically balanced study would further strengthen the human reference. We
release both the interactive labeling interface (hosted live at
\url{https://m2sv.yosubshin.com}) and the collected labels, including per-item
response-time metadata, to facilitate such larger-scale human studies and
alternative analyses of spatial reasoning difficulty.

Our transfer experiments are limited in scope and focus on generalization to a
single adjacent benchmark (MindCube). Other relevant benchmarks were not
incorporated into our initial experimental setup, either because they were not
publicly available at the time (e.g., CityCube) or were released shortly before
our experiments and follow different evaluation protocols (e.g., Ego3D-Bench,
UrBench). Extending transfer evaluations to additional spatial reasoning and
cross-view benchmarks remains an important direction for future work.

\section{Conclusion}

m2sv provides a scalable benchmark for evaluating map-to-street-view spatial
reasoning under real-world conditions. Despite strong performance on existing
multimodal benchmarks, vision–language models remain far below human accuracy on
this task, with failures increasing sharply under geometric ambiguity and
symmetry. Our analyses show that these errors are not solely due to missing
knowledge, but to difficulties in maintaining consistent allocentric–egocentric
representations and reasoning under uncertainty. By isolating this core alignment
primitive and characterizing both difficulty and failure modes, m2sv serves as a
diagnostic complement to existing spatial benchmarks and highlights open
challenges for grounded spatial reasoning.

\section*{Acknowledgments}
\label{sec:acknowledgments}

This work was supported by the National Science Foundation NRT-AI 2244574.

This work used cloud GPU resources at NCSA Delta cluster through allocation number
CIS240027 from the Advanced Cyberinfrastructure Coordination Ecosystem: Services
\& Support (ACCESS) program, which is supported by National Science Foundation
grants \#~2138259, \#~2138286, \#~2138307, \#~2137603, and \#~2138296.

The technical support and advanced computing resources from University of Hawaii
Information Technology Services -- Research Cyberinfrastructure, funded in part by
the National Science Foundation CC* awards \#~2201428 and \#~2232862 are gratefully
acknowledged.

This material is based upon work supported by the National Science Foundation CISE
Graduate Fellowships under Grant \#~2313998. Any opinions, findings, and
conclusions or recommendations expressed in this material are those of the
author(s) and do not necessarily reflect the views of the National Science
Foundation.

\FloatBarrier

\newpage 

\bibliographystyle{icml2026}
\bibliography{m2sv_paper}

\newpage 
\newpage

\appendix
\section{Background and Related Benchmarks}
\label{sec:background}

This appendix situates \textbf{m2sv} among recent benchmarks for spatial
reasoning and cross-view understanding. We focus on the benchmarks most adjacent
to \textbf{m2sv} and clarify how our task differs from them.

\subsection{What Spatial Capability Does m2sv Measure?}
\label{sec:background-capability}

\textbf{m2sv} isolates a single spatial reasoning primitive:
\emph{inferring camera heading by aligning an egocentric street-level view with a
north-up overhead map at a known location}. Each example fixes the geographic
position and varies only the candidate viewing direction, converting orientation
from a nuisance variable into the primary reasoning target. This formulation
removes global localization and retrieval from the task, forcing explicit
reasoning about road topology, landmark placement, and allocentric--egocentric
transformations under real-world ambiguity.

\subsection{Closest Related Benchmarks}
\label{sec:background-closest}

\paragraph{CityCube.}
CityCube \citep{citycube2026} benchmarks cross-view spatial reasoning in urban environments
and reports recurring failures such as directional confusion and viewpoint
inconsistency.
Compared to CityCube's broader suite of tasks, \textbf{m2sv} intentionally
constrains the problem to a \emph{single-step heading decision at a fixed
location}, enabling controlled measurement of how structural ambiguity (e.g.,
symmetry) affects orientation inference.

\paragraph{UrBench.}
UrBench \citep{urbench2024} is a multi-view urban benchmark covering diverse task types
(including tasks related to localization and orientation in urban scenes).
While UrBench contains cross-view components that are conceptually adjacent to
\textbf{m2sv}, its scope spans multiple urban reasoning dimensions.
In contrast, \textbf{m2sv} focuses narrowly on \emph{heading inference} by
aligning an overhead map with a Street View observation at a known intersection,
thereby isolating allocentric--egocentric alignment as the primary bottleneck.

\paragraph{MindCube.}
MindCube \citep{yin2025spatialmental} evaluates whether VLMs can maintain coherent spatial mental
models from limited observations, including perspective-taking and consistency
under viewpoint changes. It motivates ``map-then-reason'' style approaches that
explicitly produce intermediate spatial representations before answering.
Although m2sv shares the theme of viewpoint-consistent spatial reasoning,
\textbf{m2sv} differs in its \emph{real-world cross-view} setting (overhead map
$\leftrightarrow$ street-level imagery) and its emphasis on resolving ambiguous cross-view alignment by escalating
from coarse, unreliable cues to finer-grained spatial evidence at a concrete
intersection.

\paragraph{Ego3D-Bench.}
Ego3D-Bench \citep{ego3dbench2025} evaluates outdoor spatial reasoning using ego-centric,
multi-view observations and shows a substantial human--model gap. The associated
Ego3D-VLM framework improves performance by introducing an explicit
cognitive-map style intermediate representation derived from estimated 3D
coordinates. These results support the broader hypothesis that structured
intermediate spatial state can benefit VLM reasoning; however, whether similar
representations help \textbf{m2sv} depends on how well they align to the
map--street-view transformation and the uncertainty/ambiguity present in real
Street View observations.

\subsection{Relation to Cross-View Geo-Localization (CVGL)}
\label{sec:background-cvgl}

Cross-view geo-localization (CVGL) studies the correspondence between ground-level imagery
(e.g., street-view panoramas or limited-FoV views) and overhead imagery (e.g., satellite/aerial),
with the dominant formulation cast as \emph{retrieval}: a ground query is matched against a
geo-tagged overhead database and evaluated using retrieval metrics such as Recall@K
(e.g., on CVUSA/CVACT-style benchmarks and follow-ups)~\cite{workman2015cvusa,liu2019lending,zhu2021vigor,huang2024cvcities}.
While some benchmarks extend viewpoints~\cite{zheng2020university1652},
the core objective in these settings remains location retrieval rather than isolating directional reasoning.

Because cross-view appearance and geometry differ substantially, many CVGL methods incorporate
mechanisms to account for unknown \emph{orientation/alignment} between ground and overhead views.
Representative approaches include explicitly encoding orientation cues~\cite{liu2019lending},
using polar transforms and correlation-style matching to estimate azimuth alignment during localization~\cite{shi2020wherelooking},
or estimating cross-view alignment/orientation as part of the matching pipeline~\cite{zhu2021revisiting}.
Recent work also explores graph-structured urban priors and bearing-aware filtering on top of retrieval~\cite{shore2025spagbol},
and orientation-free formulations for cross-view matching have been proposed~\cite{hu2025oriloc}.
However, even when orientation is modeled or estimated, it is typically \emph{in service of improving retrieval},
i.e., heading is not the primary target variable.

\textbf{m2sv} complements CVGL by \emph{fixing the location} and asking the model to infer
\emph{orientation directly}. This inversion removes global retrieval shortcuts (e.g., relying on distinctive
place identity in a large gallery) and makes directional reasoning the dominant challenge.

\subsection{Structural Difficulty and Symmetry}
\label{sec:prior-difficulty}

Unlike many recent vision--language benchmarks, which define difficulty primarily
through semantic complexity or multi-step reasoning, spatial reasoning difficulty
has a long history of being operationalized through geometric transformation.
In cognitive psychology, difficulty is closely linked to mental effort and response
time (RT): classic results show that RT increases linearly with the angular
displacement required to align two objects, suggesting a process akin to physical
mental rotation~\citep{shepard1971mental}.
By contrast, contemporary VLM benchmarks often categorize difficulty by reasoning
depth or semantic richness (e.g., number of symbols, hops, or compositional steps),
as in MathVista~\citep{lu2024mathvista} or MapQA~\citep{li2025mapqa}, potentially
underrepresenting the role of intrinsic geometric ambiguity.

\textbf{m2sv} introduces \emph{structural difficulty} rooted in intersection
geometry and symmetry. Symmetric or near-symmetric intersections induce multiple
plausible orientations at a coarse level, creating ``symmetry traps'' that cannot
be resolved by semantic pattern matching alone.
Breaking these traps requires escalation to finer-grained spatial cues
(e.g., lane markings, curb geometry, subtle landmark placement), a process that
mirrors increased human RT in geometric alignment tasks.
This framing connects classic findings on mental rotation
to modern observations of allocentric--egocentric failures in VLMs,
and isolates a form of difficulty that is intrinsic to the scene structure itself.

\begin{table}[t]
  \centering
  \caption{Summary statistics of the m2sv-20k dataset.}
  \label{tab:dataset-stats}
  \begin{tabular}{l r}
    \toprule
    Statistic & Value \\
    \midrule
    Number of examples & 20,000 \\
    Number of cities & 32 \\
    Candidate directions (median) & 3 \\
    Street View distance (median) & 2.05 m \\
    Street View distance (95\%) & 4.51 m \\
    \bottomrule
  \end{tabular}
\end{table}

\begin{table}[t]
  \centering
  \caption{Cross-benchmark evaluation for Qwen3-VL-8B variants.}
  \label{tab:transfer}
  \begin{tabular}{l c c c c c}
    \toprule
    Model & MMMU & MME-Cog/Per & MMStar & Omni3D \\
    \midrule
    Base & 0.556 & 643 / 1721 & 0.604 & 0.388 \\
    SFT & 0.574 & 501 / 1525 & 0.574 & 0.372 \\
    SFT+RL & 0.563 & 550 / 1670 & 0.644 & -- \\
    \bottomrule
  \end{tabular}
\end{table}

\begin{table}[t]
  \centering
  \caption{MindCube evaluation suite with varying input/output formats.}
  \label{tab:m2sv-generalization}
  \begin{tabular}{l l c c c}
    \toprule
    Input & Output Format & Base & SFT & SFT+RL \\
    \midrule
    Raw + q & Aug + rsn $\rightarrow$ ans & 34.5 & 34.0 & 33.2 \\
    Aug. + QA & Direct answer & 24.6 & 37.6 & 41.0 \\
    Aug. + QA & Rsn $\rightarrow$ answer & 46.9 & 40.3 & 46.1 \\
    Raw + q & Rsn $\rightarrow$ answer & 38.5 & 36.8 & 36.6 \\
    Raw + q & Map + rsn $\rightarrow$ ans & 34.7 & 33.5 & 35.2 \\
    Raw + q & Direct answer & 27.0 & 34.9 & 34.7 \\
    \bottomrule
  \end{tabular}
\end{table}

\begin{figure*}[t]
  \centering
  \renewcommand{\arraystretch}{1.15}
  
  \begin{minipage}[t]{\linewidth}
    \centering
    \begin{minipage}[t]{0.24\linewidth}
      \centering
      \vspace{0pt}
      \includegraphics[width=\linewidth]{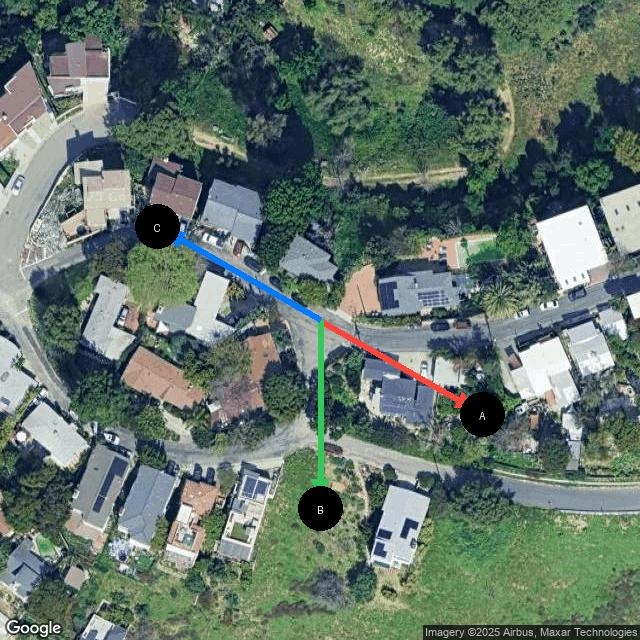}
    \end{minipage}
    \hfill
    \begin{minipage}[t]{0.24\linewidth}
      \centering
      \vspace{0pt}
      \includegraphics[width=\linewidth]{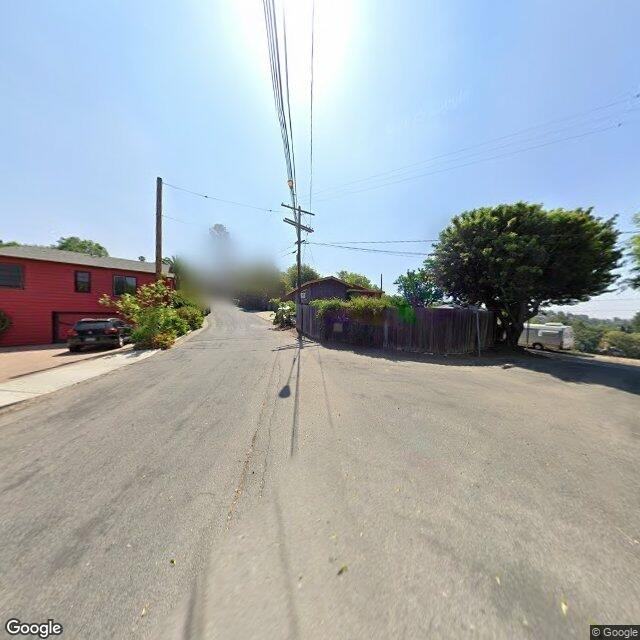}
    \end{minipage}
    \hfill
    \begin{minipage}[t]{0.50\linewidth}
      \vspace{0pt}
      \fbox{
        \begin{minipage}[t]{\linewidth}
        \footnotesize
        \textbf{Ground truth:} A, \textbf{Prediction:} C
        \medskip

        \textbf{Model trace (excerpt):}
  
        Looking from the center towards A, the road curves to the right. This contradicts the left curve seen in the photo.
  
        \medskip
        \textbf{Failure:}
        Egocentric--allocentric left/right inversion: Overhead map shows a left curve, but model reasons as if it were a right curve.
        \end{minipage}
      }
    \end{minipage}
  
    \vspace{0.4em}
    \centering
    {\small\textbf{(a) Egocentric--allocentric confusion (left--right inversion)}}
  \end{minipage}
  
  \vspace{0.8em}
  
  \begin{minipage}[t]{\linewidth}
    \centering
    \begin{minipage}[t]{0.24\linewidth}
      \centering
      \vspace{0pt}
      \includegraphics[width=\linewidth]{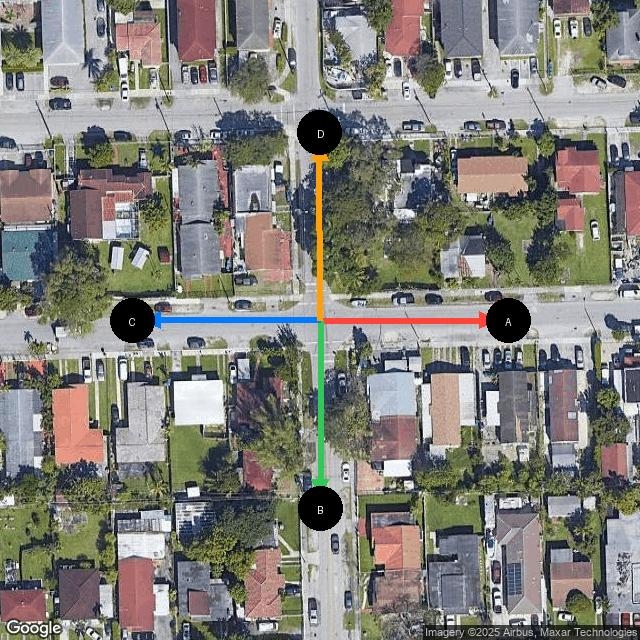}
    \end{minipage}
    \hfill
    \begin{minipage}[t]{0.24\linewidth}
      \centering
      \vspace{0pt}
      \includegraphics[width=\linewidth]{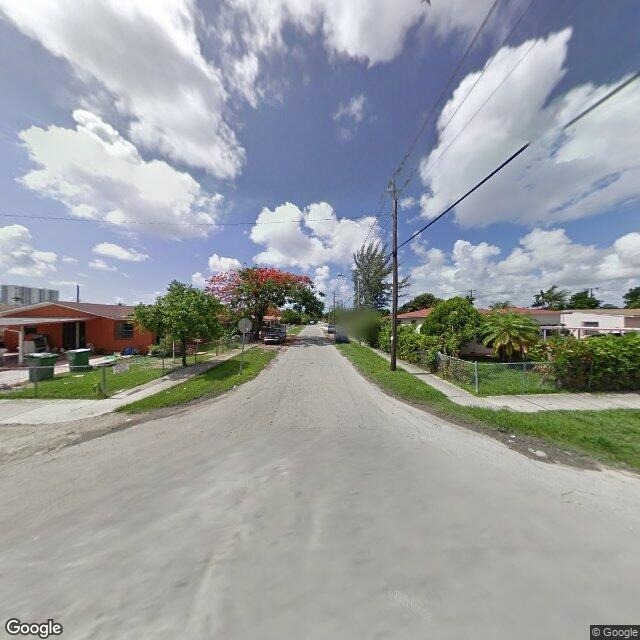}
    \end{minipage}
    \hfill
    \begin{minipage}[t]{0.50\linewidth}
      \vspace{0pt}
      \fbox{
        \begin{minipage}[t]{\linewidth}
        \footnotesize
        \textbf{Ground truth:} B, \textbf{Prediction:} D
        \medskip

        \textbf{Model trace (excerpt):}
  
        Looking south from the center, the buildings on both corners appear to have gray roofs. There is no prominent reddish-orange house on the left corner as seen in the photo.
Therefore, B is incorrect.
  
        \medskip
        \textbf{Failure:}
        The Street View angle of the roof is shallow, and the roof color is lighting-dependent.
        Humans would discount this cue.

        \end{minipage}
      }
    \end{minipage}
  
    \vspace{0.4em}
    \centering
    {\small\textbf{(b) Over-reliance on unreliable visual cues}}
  \end{minipage}

  \vspace{0.8em}
  
  \begin{minipage}[t]{\linewidth}
    \centering
    \begin{minipage}[t]{0.24\linewidth}
      \centering
      \vspace{0pt}
      \includegraphics[width=\linewidth]{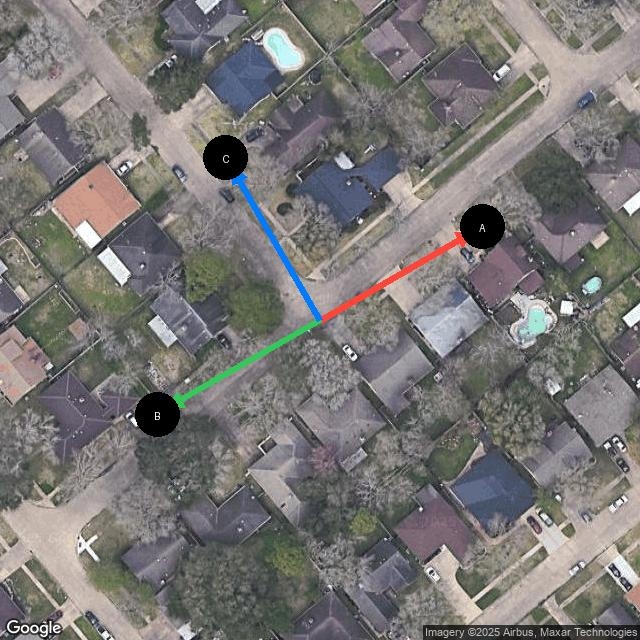}
    \end{minipage}
    \hfill
    \begin{minipage}[t]{0.24\linewidth}
      \centering
      \vspace{0pt}
      \includegraphics[width=\linewidth]{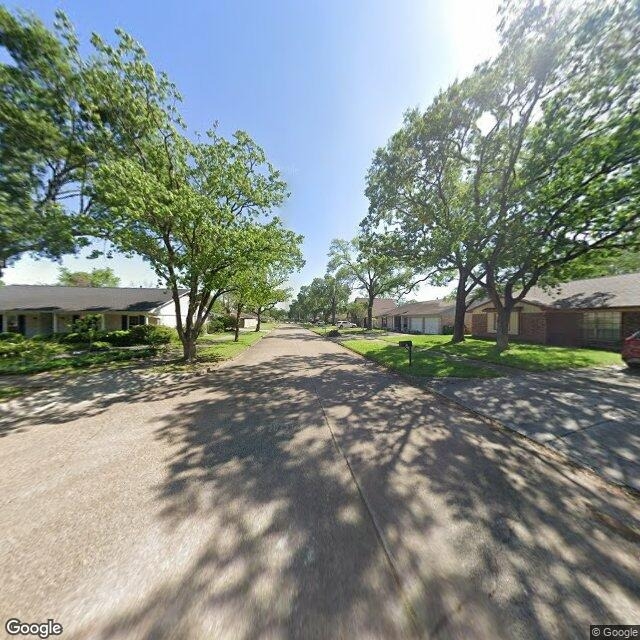}
    \end{minipage}
    \hfill
    \begin{minipage}[t]{0.50\linewidth}
      \vspace{0pt}
      \fbox{
        \begin{minipage}[t]{\linewidth}
        \footnotesize
        \textbf{Ground truth:} A, \textbf{Prediction:} B
        \medskip

        \textbf{Model trace (excerpt):}
  
        Facing direction A (east), we look down a relatively straight road.
        To the immediate left (north) of this view is a corner lot. The house on this corner has a large swimming pool in its backyard. This does not match the houses seen in the street view, which do not have prominent pools immediately adjacent to the road like this one does.
  
        \medskip
        \textbf{Failure:}
        The corner house does not have a pool, but a pool exists two houses down.
        The model misattributes a nearby landmark to the wrong spatial anchor.
        \end{minipage}
      }
    \end{minipage}
  
    \vspace{0.4em}
    \centering
    {\small\textbf{(c) Landmark misbinding}}
  \end{minipage}

  \caption{
  \textbf{Representative qualitative failure modes on m2sv.}
  Each row shows the same map--street-view pair with a different erroneous reasoning
  pattern exhibited by the model. From left to right: overhead map with candidate
  directions, corresponding street-view image, and an excerpt from the model’s
  reasoning trace highlighting the incorrect assumption. All examples are correctly
  solved by human annotators.
  }
  \label{fig:failure-examples}
  \end{figure*}

\begin{figure*}[t]
  \centering
  \renewcommand{\arraystretch}{1.15}
  
  \begin{minipage}[t]{\linewidth}
    \centering
    \begin{minipage}[t]{0.24\linewidth}
      \centering
      \vspace{0pt}
      \includegraphics[width=\linewidth]{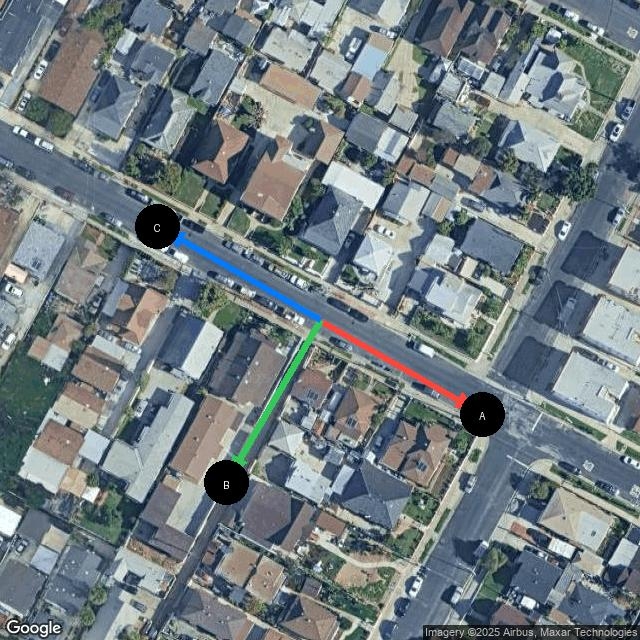}
    \end{minipage}
    \hfill
    \begin{minipage}[t]{0.24\linewidth}
      \centering
      \vspace{0pt}
      \includegraphics[width=\linewidth]{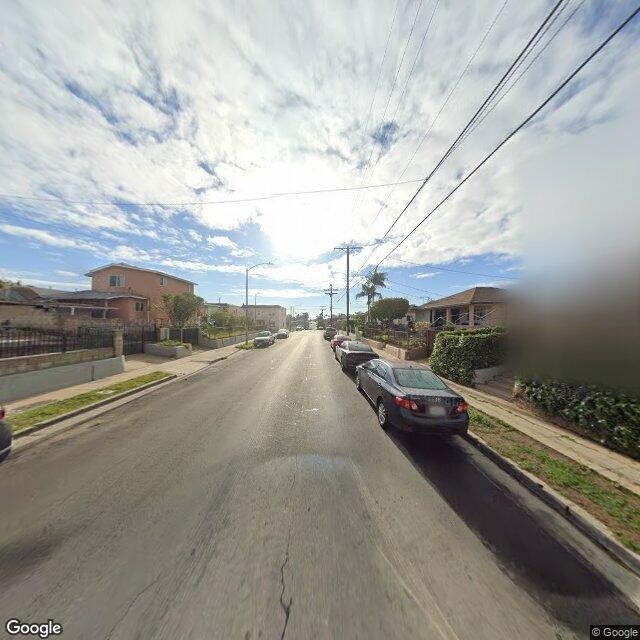}
    \end{minipage}
    \hfill
    \begin{minipage}[t]{0.50\linewidth}
      \vspace{0pt}
      \fbox{
        \begin{minipage}[t]{\linewidth}
        \footnotesize
        \textbf{Ground truth:} A, \textbf{Prediction:} C
        \medskip

        \textbf{Model trace (excerpt):}
  
        Direction A: This arrow points east-southeast down a relatively narrow residential street. It has no crosswalks visible across its width. \\
        Direction C: This arrow points west-northwest down a much wider road than A and B. This main road clearly has multiple lanes separated by what appear to be painted lines, indicating it's a major thoroughfare. Crosswalks are visible crossing this road.
  
        \medskip
        \textbf{Failure:}
        A and C are two sides of the same street so the road width is the same, but the answer emphasizes the road width as the key factor
        \end{minipage}
      }
    \end{minipage}
  
    \vspace{0.4em}
    \centering
    {\small\textbf{(a) Failure to maintain a consistent spatial representation}}
  \end{minipage}

  \vspace{0.8em}

  \begin{minipage}[t]{\linewidth}
    \centering
    \begin{minipage}[t]{0.24\linewidth}
      \centering
      \vspace{0pt}
      \includegraphics[width=\linewidth]{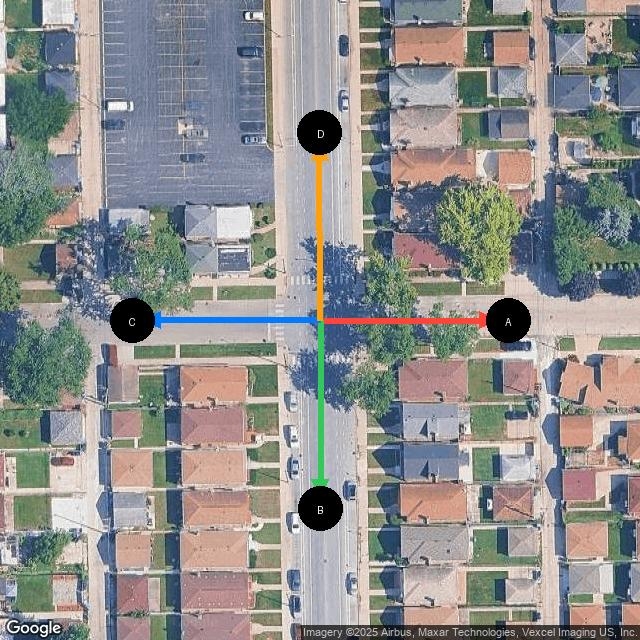}
    \end{minipage}
    \hfill
    \begin{minipage}[t]{0.24\linewidth}
      \centering
      \vspace{0pt}
      \includegraphics[width=\linewidth]{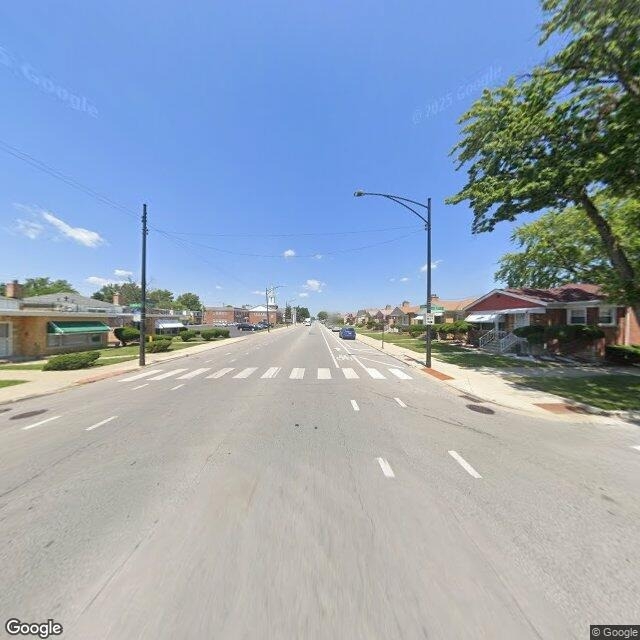}
    \end{minipage}
    \hfill
    \begin{minipage}[t]{0.50\linewidth}
      \vspace{0pt}
      \fbox{
        \begin{minipage}[t]{\linewidth}
        \footnotesize
        \textbf{Ground truth:} D, \textbf{Prediction:} C
        \medskip

        \textbf{Model trace (excerpt):}
  
        Option D: This view looks up the main, wide road. On the northeast corner (on the right), there is a residential house. On the northwest corner (on the left), there is the commercial building. This arrangement is the opposite of what is seen in the street view photo, where the commercial building is on the left and the house is on the right.
  
        \medskip
        \textbf{Failure:}
        The model correctly identifies left–right relationships but subsequently contradicts its own assignments, losing track of spatial state during verbalization.
        \end{minipage}
      }
    \end{minipage}
  
    \vspace{0.4em}
    \centering
    {\small\textbf{(b) Internal contradiction and state inconsistency}}
  \end{minipage}

  \vspace{0.8em}

  \begin{minipage}[t]{\linewidth}
    \centering
    \begin{minipage}[t]{0.24\linewidth}
      \centering
      \vspace{0pt}
      \includegraphics[width=\linewidth]{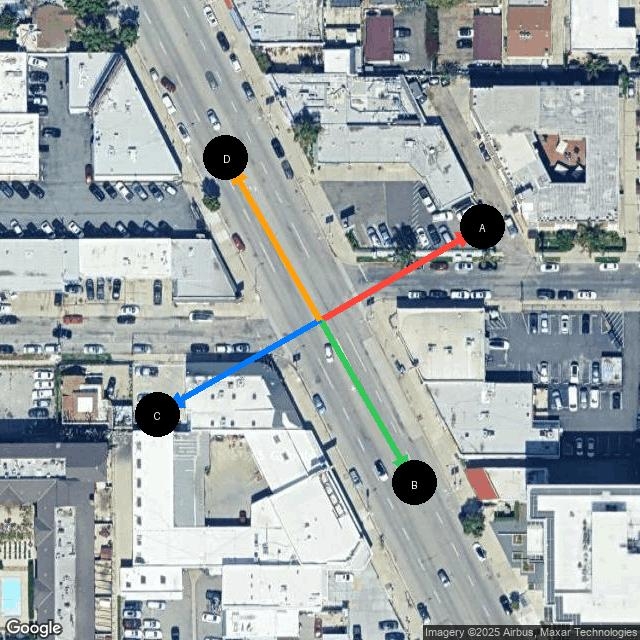}
    \end{minipage}
    \hfill
    \begin{minipage}[t]{0.24\linewidth}
      \centering
      \vspace{0pt}
      \includegraphics[width=\linewidth]{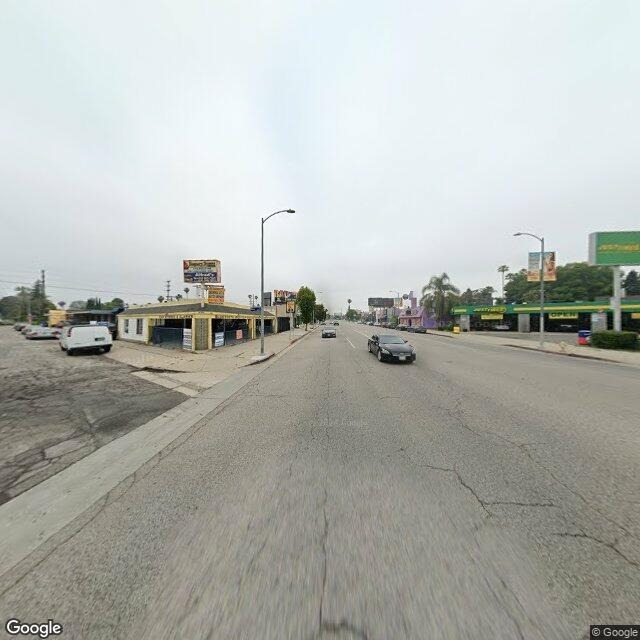}
    \end{minipage}
    \hfill
    \begin{minipage}[t]{0.50\linewidth}
      \vspace{0pt}
      \fbox{
        \begin{minipage}[t]{\linewidth}
        \footnotesize
        \textbf{Ground truth:} D, \textbf{Prediction:} B
        \medskip

        \textbf{Model trace (excerpt):}
  
        For view B: Left = Building A, Right = Long Rectangular Building.
        For view D: Left = Large Building with Parking Lot, Right = Building A.
        Comparing the photo to view B: We see a specific yellowish building on the left and a gas station on the right.
  
        \medskip
        \textbf{Failure:}
        The model correctly narrows the decision to two symmetric candidates (B and D),
but fails to use subtle geometric cues—such as building orientation relative to
the road—to break symmetry, instead introducing unsupported semantic evidence
(e.g., a gas station).
        \end{minipage}
      }
    \end{minipage}
  
    \vspace{0.4em}
    \centering
    {\small\textbf{(c) Symmetry traps and insufficient cue escalation.}}
  \end{minipage}

  \vspace{0.8em}
  
  \begin{minipage}[t]{\linewidth}
    \centering
    \begin{minipage}[t]{0.24\linewidth}
      \centering
      \vspace{0pt}
      \includegraphics[width=\linewidth]{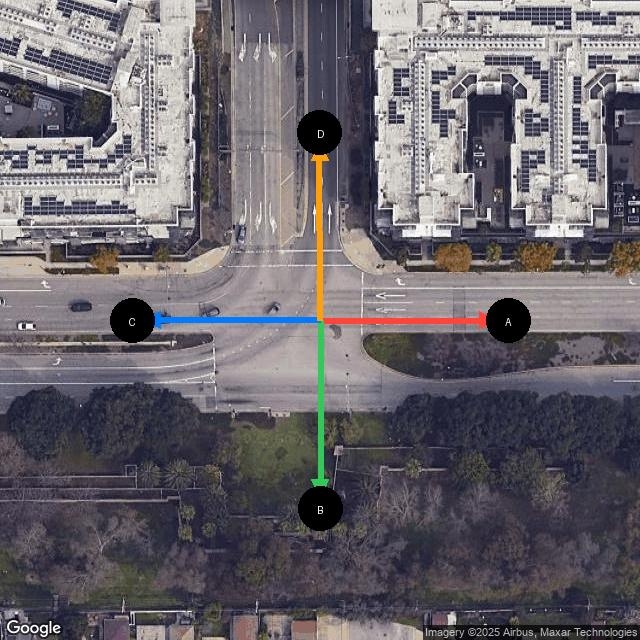}
    \end{minipage}
    \hfill
    \begin{minipage}[t]{0.24\linewidth}
      \centering
      \vspace{0pt}
      \includegraphics[width=\linewidth]{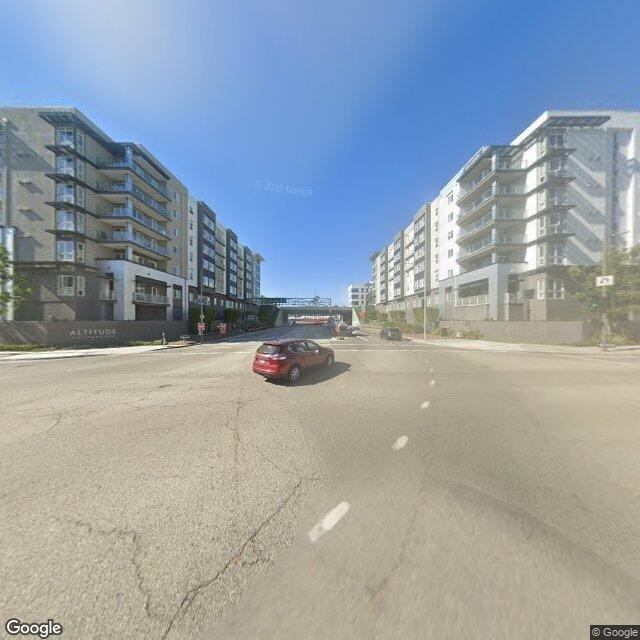}
    \end{minipage}
    \hfill
    \begin{minipage}[t]{0.50\linewidth}
      \vspace{0pt}
      \fbox{
        \begin{minipage}[t]{\linewidth}
        \footnotesize
        \textbf{Ground truth:} D, \textbf{Prediction:} A
        \medskip

        \textbf{Model trace (excerpt):}

        Direction A provides the most complete match. It correctly places large buildings on both sides of the road and crucially includes the distant underpass/bridge structure that is clearly visible in the street view photo.

        \medskip
        \textbf{Failure:}
        The model invents or misattributes a nonexistent large-scale structure (underpass/bridge) and then reasons as if it were present in the map.
        \end{minipage}
      }
    \end{minipage}
  
    \vspace{0.4em}
    \centering
    {\small\textbf{(d) False landmark projection}}
  \end{minipage}

  \vspace{0.5em}
  
  \caption{
  \textbf{More qualitative failure modes on m2sv.}
  Each row shows the same map--street-view pair with a different erroneous reasoning
  pattern exhibited by the model. From left to right: overhead map with candidate
  directions, corresponding street-view image, and an excerpt from the model’s
  reasoning trace highlighting the incorrect assumption. All examples are correctly
  solved by human annotators.
  }
  \label{fig:failure-examples-appendix}
  \end{figure*}

\end{document}